\documentclass[11pt]{article}

\usepackage{acl}
\usepackage{times}
\usepackage{latexsym}
\usepackage[T1]{fontenc}
\usepackage[utf8]{inputenc}
\DeclareUnicodeCharacter{266B}{}
\usepackage{microtype}
\usepackage{graphicx}
\usepackage{booktabs}
\usepackage{amsmath}
\usepackage{amssymb}
\usepackage{url}
\usepackage{xcolor}
\usepackage{multirow}
\usepackage{array}
\usepackage{enumitem}
\usepackage{subcaption}
\usepackage{tikz}
\usepackage{tabularx}
\usepackage{placeins}
\usepackage{listings}
\usetikzlibrary{shapes.geometric, arrows.meta, positioning, fit, calc}

\lstdefinestyle{ragustyle}{
  basicstyle=\ttfamily\footnotesize,
  breaklines=true,
  showstringspaces=false,
  keywordstyle=\color{blue!70!black},
  commentstyle=\color{green!50!black},
  language=Python,
  numbers=none,
  frame=single,
  framerule=0.3pt,
  rulecolor=\color{black!30},
  xleftmargin=4pt,
  xrightmargin=4pt,
}
\newcommand{\ragu}{RAGU}
\newcommand{\menolite}{Meno-Lite-0.1}
\newcommand{\hipporoot}{https://github.com/OSU-NLP-Group/HippoRAG/blob/d437bfb1805278b81e20c82357ed3f7d90f14901/}
\newcommand{\nereltask}{the \texttt{nerel-bench} task group}

\title{RAGU: A Multi-Step GraphRAG Engine\\with a Compact Domain-Adapted LLM}

\author{
  \textbf{Mikhail Komarov}$^{1,*}$ \quad
  \textbf{Ivan Bondarenko}$^{2,*}$ \quad
  \textbf{Stanislav Shtuka}$^{1,3}$ \quad
  \textbf{Oleg Sedukhin}$^{\dagger}$ \\
  \textbf{Roman Shuvalov}$^{1}$ \quad
  \textbf{Yana Dementyeva}$^{2}$ \quad
  \textbf{Matvey Solovyov}$^{1}$ \quad
  \textbf{Nikolay O. Nikitin}$^{1}$ \\[4pt]
  $^{1}$ITMO University \quad
  $^{2}$Novosibirsk State University \quad
  $^{3}$Far Eastern Federal University \\
  $^{\dagger}$Independent Researcher \quad
  $^{*}$Equal contribution \\
  \texttt{i.bondarenko@g.nsu.ru}
}

\date{}

\begin{document}

\maketitle

\begin{abstract}
Graph retrieval-augmented generation (GraphRAG) enhances large language
models with structured knowledge, yet existing systems construct knowledge
graphs in a single extraction pass, producing noisy entities and brittle
retrieval. \ragu{}, an open-source modular GraphRAG engine, addresses this
by separating extraction from consolidation: entities and relations pass
through two-stage typed extraction, DBSCAN-backed deduplication, LLM
summarization, and Leiden community detection. A key insight motivates a
compact extractor: the skills an in-pipeline LLM needs---comprehension,
extraction, reasoning over context---are language skills that grow only
weakly with model size, unlike factual world knowledge. Accordingly, we
train \menolite{}, a 7\,B model optimized for language skills, which
outperforms Qwen2.5-32B on knowledge-graph construction (+12.5\% relative
harmonic mean) and matches it on English GraphRAG tasks. On GraphRAG-Bench
(Medical), \ragu{} retrieves the most complete context at every factoid
level (evidence recall up to 0.84 vs.\ $\leq$0.76) and overtakes
HippoRAG\,2 on synthesis tasks; on multi-hop factoid QA, the apparent
HippoRAG\,2 advantage is shown to be largely an answer-format artifact.
\ragu{} is installable via \texttt{pip install graph\_ragu}, runs on a
single GPU, and is released under MIT license. The source code is publicly available at \url{https://github.com/RaguTeam/RAGU},
and the \menolite{} model can be obtained from
\url{https://huggingface.co/bond005/meno-lite-0.1}.
\end{abstract}

\section{Introduction}
\label{sec:intro}

Retrieval-augmented generation~(RAG) grounds large language models (LLMs)
in external knowledge~\cite{lewis2020rag,gao2023rag_survey}. Traditional
RAG retrieves flat chunks without capturing cross-document entity
relationships. Graph RAG~(GraphRAG)~\cite{edge2024graphrag,guo2025lightrag,gutierrez2025hipporag2}
addresses this by building a knowledge graph and using graph traversal
during retrieval, but practical adoption faces three obstacles.

\paragraph{Obstacle 1: Single-pass extraction.}
Current systems treat knowledge graph construction as a single LLM
extraction pass, producing noisy, duplicated entities with no mechanism
to \emph{consolidate} information across chunks.

\paragraph{Obstacle 2: Dependence on expensive LLMs.}
Extraction quality determines graph quality, so practitioners default to
large API models (GPT-4-class). This rests on a false premise: the
capabilities an LLM needs \emph{inside} a RAG pipeline---comprehension,
extraction, reasoning over context---are \emph{language skills}, not
factual recall. As we show next, language skills grow weakly with model
size, while world knowledge scales steeply. A compact, skill-oriented
model is therefore sufficient.

\paragraph{Obstacle 3: Engineering immaturity.}
Many open-source frameworks suffer from installation failures or unsafe
code paths (e.g., \texttt{eval()} on raw LLM output). A GraphRAG engine
should be both \emph{semantically} strong and \emph{engineerable}:
installable, testable, and deployable on cost-effective hardware.

\begin{figure}[t]
\centering
\includegraphics[width=\columnwidth]{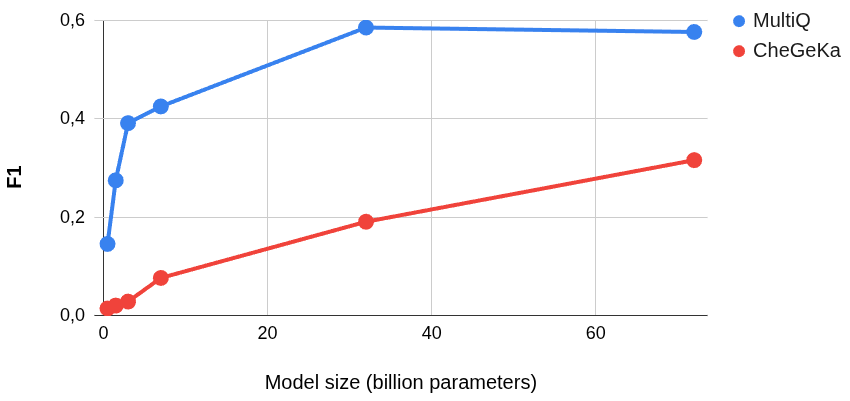}
\caption{Effect of model size on world-knowledge (CheGeKa) vs.\
  language-skill (MultiQ) tasks in the Qwen2.5-Instruct family
  (F1 scores on MERA~\cite{fenogenova2024mera}). CheGeKa F1 grows
  21.1$\times$ from 0.5\,B to 72\,B; MultiQ only 4$\times$. Log-linear
  slopes: 0.65 vs.\ 0.26.}
\label{fig:hypothesis}
\end{figure}

\paragraph{Language/World Knowledge Hypothesis.}
We hypothesize that \textbf{world knowledge} scales near-linearly with
parameter count, whereas \textbf{language skills} scale markedly more
slowly. Figure~\ref{fig:hypothesis} tests this on the Qwen2.5-Instruct
family~\cite{yang2024qwen2}: on CheGeKa~\cite{taktasheva-etal-2022-tape}
(world-knowledge quiz), F1 increases 21.1$\times$ from 0.5\,B to 72\,B,
while on MultiQ (all facts in-context), it increases only 4$\times$.
In a GraphRAG pipeline, the LLM extracts entities, summarizes
descriptions, and generates answers from context---all language skills.
This prediction motivates a compact extractor.

We address all three obstacles with two artifacts, released under open licenses:
\begin{enumerate}[nosep,leftmargin=*]
  \item \textbf{\menolite{}}, a 7\,B model fine-tuned from
        RuadaptQwen2.5-7B~\cite{tikhomirov2025ruadapt} for RAG-oriented
        language skills over the NEREL schema~\cite{loukachevitch2021nerel}.
        It outperforms Qwen2.5-32B on KG construction (+12.5\% harmonic
        mean) and rivals models up to $\sim$4$\times$ larger.
  \item \textbf{\ragu{}}, a modular multi-step GraphRAG engine whose
        pipeline separates extraction from consolidation---two-stage typed
        extraction, DBSCAN-backed summarization, Leiden community
        detection---yielding cleaner and more connected knowledge graphs.
        It is installable via \texttt{pip install graph\_ragu} and runs on
        a single GPU.
\end{enumerate}

\noindent
\ragu{} differs from prior systems---Microsoft
GraphRAG~\cite{edge2024graphrag}, LightRAG~\cite{guo2025lightrag},
HippoRAG\,2~\cite{gutierrez2025hipporag2},
Wikontic~\cite{chepurova2026wikontic}---by introducing an explicit
multi-step consolidation stage and by targeting engineering maturity.

\section{System Description}
\label{sec:system}

\begin{figure*}[!t]
  \centering
  \includegraphics[width=\textwidth]{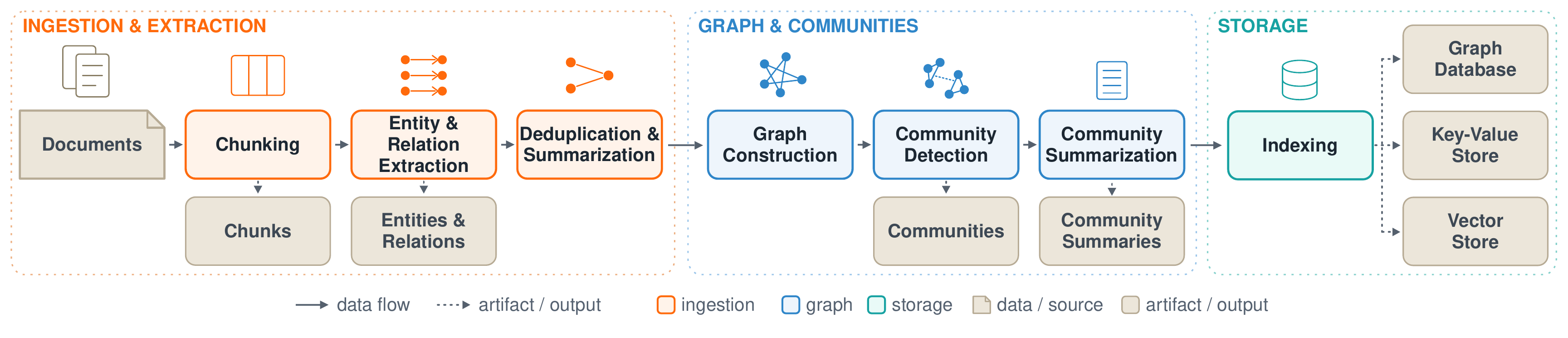}
  \caption{End-to-end indexing pipeline. Documents are chunked, entities
    and relations are extracted under the NEREL schema, deduplicated and
    summarized, then grouped into communities via Leiden clustering.
    All artifacts persist across three swappable storage tiers (graph
    database, key-value store, vector store). Solid arrows indicate data
    flow between pipeline stages; dashed arrows indicate artifacts
    produced at each stage.}
  \label{fig:architecture}
\end{figure*}

\subsection{Multi-Step Graph Construction}

\ragu{} processes documents through six configurable stages
(Figure~\ref{fig:architecture}):

\paragraph{Step 1: Chunking.}
Three strategies: \emph{SimpleChunker} (fixed-size overlapping chunks),
\emph{SemanticTextChunker} (embedding-based split points), and
\emph{SmartSemanticChunker} (cross-encoder reranking).

\paragraph{Step 2: Two-stage extraction.}
Unlike single-pass systems, \ragu{} separates entity extraction (Stage~1)
from relation extraction (Stage~2). Entities are first extracted and
validated against the NEREL schema (29 entity types, 49 relation types),
then fed back as constraints: every \texttt{source\_entity} and
\texttt{target\_entity} in a relation \emph{must} match a validated entity
name. This eliminates spurious entity--relation mismatches. Optional
in-context-learning (ICL) examples are injected at both stages, selected
by semantic, BM25, hybrid, or random strategies.

\paragraph{Step 3: Consolidation.}
\emph{EntitySummarizer} groups entities by (name, type) and---for entities
with many duplicate mentions---applies DBSCAN clustering and LLM
summarization. \emph{RelationSummarizer} follows the same pattern.
Reducing noise \emph{before} community detection produces a cleaner
graph---the step absent from single-pass systems like LightRAG.

\paragraph{Steps 4--6: Community detection, summarization, refinement.}
Hierarchical Leiden clustering partitions the deduplicated graph; an LLM
generates structured community reports (title, summary, findings);
pluggable modules (e.g., \texttt{RemoveIsolatedNodes}) optionally refine.

\subsection{Search Engines}
\label{sec:search}

\ragu{} provides five engines: \textbf{LocalSearch} (vector-similarity
entity retrieval expanded to relations and chunks), \textbf{GlobalSearch}
(LLM-rated community summarization), \textbf{NaiveSearch} (standard
vector RAG), \textbf{MixSearch} (parallel multi-engine), and
\textbf{QueryPlanEngine} (DAG decomposition). All support cross-encoder
reranking and hybrid dense+sparse retrieval via Qdrant.

\subsection{Engineering and Deployment}
\label{sec:engineering}

\ragu{} is a production-ready Python package
(see Appendix~\ref{app:engineering} for a comparison with HippoRAG\,2).
\emph{(i)}~A three-tier storage abstraction (graph/KV/vector) with
lifecycle callbacks enables backend swapping
(NetworkX$\to$Neo4j, NanoVDB$\to$Qdrant).
\emph{(ii)}~An async-first API with bounded concurrency provides
production-safe throughput.
\emph{(iii)}~Structured LLM outputs are validated through Pydantic v2,
removing manual JSON post-processing and preventing code injection.
\emph{(iv)}~Incremental upsert/update/delete with deterministic
hash-based IDs and merge policies, plus a consistency auditor verifying
cross-store integrity.
$\sim$374 tests and a deterministic mock LLM server enable CI without
API keys. The system supports single-GPU deployment with a 7\,B
extraction model.

\subsection{Compact Model: \menolite{}}
\label{sec:meno}

\menolite{}~\cite{bondarenko-etal-2026-raguteam} is derived from
RuadaptQwen2.5-7B-Lite-Beta~\cite{tikhomirov2025ruadapt} through
continued pretraining (1.3B tokens, Russian+English educational/scientific
texts) and supervised fine-tuning (50M tokens) covering NEREL-based
extraction~\cite{loukachevitch2021nerel}, multi-hop
QA~\cite{tang2024multihop,Katsis2025mtrag}, and query logs. The critical
distinction from general-purpose LLMs: instructions teach the model to
\emph{use context} rather than \emph{recall facts}---investing compute
in language skills, not world knowledge.
Key properties: 128K context window (passkey retrieval 0.98 at 128K),
47\% better tokenizer efficiency than vanilla Qwen2.5 on Russian text
(3.77 vs.\ 2.57 chars/token), and single-consumer-GPU deployment via
vLLM~\cite{kwon2023efficient}.
The model and training details are documented in the public model card
at \url{https://huggingface.co/bond005/meno-lite-0.1}.

\section{Evaluation}
\label{sec:eval}

\subsection{Setup}
\label{sec:setup}

We evaluate on four GraphRAG benchmarks:
\emph{GraphRAG-Bench}~\cite{xiang2026graphragbench} (Medical domain;
four difficulty levels: fact retrieval, complex reasoning, contextual
summarize, creative generation),
\emph{BioASQ}~\cite{BioASQ2023},
\emph{MuSiQue}~\cite{trivedi2022musique}, and
\emph{2WikiMultiHopQA}~\cite{ho2020multihopqa}.
All systems use the \emph{same} answer-generation LLM (gpt-4o-mini),
isolating graph construction quality. Graph construction LLMs are varied
as independent variables: \menolite{} (7\,B) and gpt-oss-20b or
Qwen2.5-7B as the second index LLM. Metrics include Answer Correctness
(AC; LLM-judge), ROUGE-L, Coverage, Faithfulness, Evidence Recall~(ER),
and Context Relevancy. LLM-as-judge evaluations use
\texttt{google/gemini-3-flash-preview}, ensuring no evaluator--generator
overlap.

\begin{table*}[!t]
\centering
\small
\setlength{\tabcolsep}{5pt}
\caption{Generation quality on GraphRAG-Bench (Medical domain).
  AC\,=\,Answer Correctness, Cov\,=\,Coverage,
  Faith\,=\,Faithfulness (all $\times$100).
  All systems use bge-large-en-v1.5 for embeddings and gpt-4o-mini for
  answer generation; only the graph-construction LLM varies.
  \ragu{} rows use ICL\,=\,1 and Val\,=\,yes.
  Best per column in \textbf{bold}.}
\label{tab:gen_medical}
\begin{tabular}{l l c c cc ccc}
\toprule
& & \multicolumn{1}{c}{Fact Retr.}
& \multicolumn{1}{c}{Complex Reas.}
& \multicolumn{2}{c}{Contextual Summ.}
& \multicolumn{3}{c}{Creative Gen.} \\
\cmidrule(lr){3-3}\cmidrule(lr){4-4}\cmidrule(lr){5-6}\cmidrule(lr){7-9}
System & Index LLM & AC & AC & AC & Cov & AC & Cov & Faith \\
\midrule
LightRAG    & Qwen2.5-7B  & 25.9 & 20.3 & 22.1 & 51.0 & 14.2 & 3.1 & 23.1 \\
LightRAG    & \menolite{} & 26.2 & 20.2 & 22.6 & 51.2 & 14.4 & 3.9 & 27.6 \\
\midrule
HippoRAG\,2 & Qwen2.5-7B  & \textbf{72.7} & 67.9 & 64.6 & 51.6 & 57.2 & 33.2 & 31.5 \\
HippoRAG\,2 & \menolite{} & 72.4 & \textbf{68.4} & \textbf{65.0} & 51.7 & 56.9 & 34.7 & 26.6 \\
\midrule
\ragu{}     & Qwen2.5-7B  & 54.1 & 54.6 & 64.9 & \textbf{73.2} & 58.1 & 56.4 & \textbf{35.1} \\
\ragu{}     & \menolite{} & 54.2 & 53.7 & 64.1 & 71.1 & \textbf{59.0} & \textbf{57.4} & 34.2 \\
\bottomrule
\end{tabular}
\end{table*}

\subsection{GraphRAG-Bench Results}
\label{sec:graphragbench}

\begin{figure*}[!t]
\centering
\includegraphics[width=\textwidth]{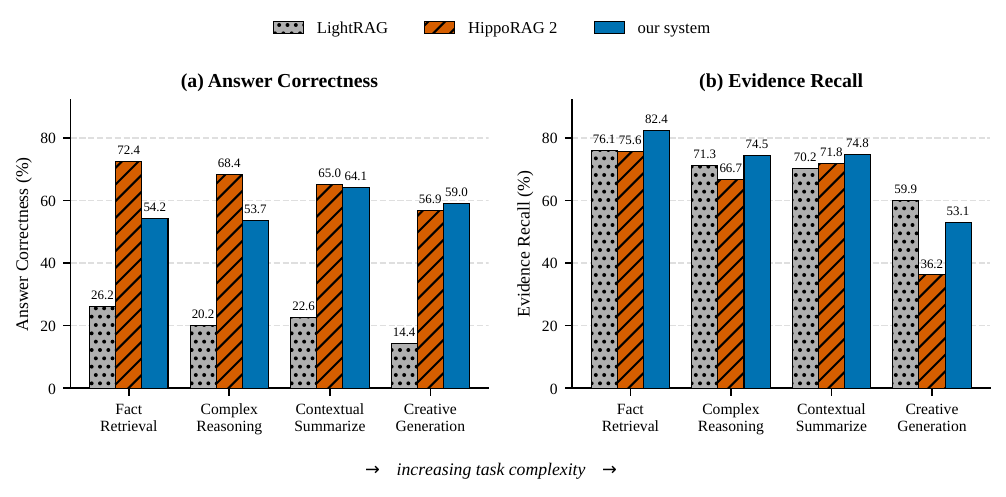}
\caption{Cross-over by task complexity on GraphRAG-Bench (Medical). All
  three systems use \menolite{} (7\,B) as the index LLM and gpt-4o-mini
  for answer generation. (a)~Answer Correctness: \ragu{} trails
  HippoRAG\,2 on Fact Retrieval, but the gap closes and \ragu{} leads on
  Creative Generation. (b)~Evidence Recall: \ragu{} retrieves the most
  complete context on all factoid levels. Task difficulty increases left
  to right.}
\label{fig:crossover}
\end{figure*}

Table~\ref{tab:gen_medical} and Figure~\ref{fig:crossover} reveal a
cross-over. On the two \emph{factoid} levels, HippoRAG\,2 leads: its
personalized PageRank pins down single facts (AC 72.4 vs.\ 54.2 for
\ragu{} on Fact Retrieval, $\Delta{=}{-}18.2$\,pp). As tasks demand broad
synthesis rather than chain-following, the gap closes
monotonically---Complex Reasoning (${-}14.7$\,pp), Contextual Summarize
(${-}0.9$\,pp, parity)---and \emph{reverses} on Creative Generation,
where \ragu{} wins AC (59.0 vs.\ 56.9) and Faithfulness (34.2 vs.\ 26.6).
On Coverage, the metric that directly rewards retrieving \emph{all}
relevant material, \ragu{} leads throughout (57.4 vs.\ 34.7 on Creative
Generation). LightRAG is weakest at every level, confirming that
single-pass free-form extraction produces a structurally poorer graph
than typed multi-step consolidation.

Retrieval results confirm the mechanism: \ragu{} attains the highest
Evidence Recall at \emph{every factoid} level (84 vs.\ $\leq$76\% for
competitors), directly supporting the consolidation hypothesis.
That HippoRAG\,2 nonetheless wins factoid AC despite lower Evidence
Recall reflects the precision of its chain traversal on single-fact
queries.
Ablation results (Appendix~\ref{app:ablation}) show that the ICL and
validation toggles each shift AC by at most $\sim$1.5\,pp, so the gap to
competitors cannot be attributed to those options; it reflects the
aggregate effect of \ragu's structural choices.  Across extraction LLMs
from 3\,B to 14\,B, AC varies by at most $\sim$1.5\,pp, confirming that
model size does not affect graph quality.

\subsection{Multi-Hop QA Results}
\label{sec:multihop}

\begin{table}[!t]
\centering
\small
\caption{Multi-hop QA under two answer-generation protocols.
  AC\,=\,Answer Correctness, RL\,=\,ROUGE-L ($\times$100).
  All systems use gte-multilingual-base for embeddings and gpt-4o-mini
  for answer generation; only the graph-construction LLM varies
  (GPT\,=\,gpt-oss-20b, Ours\,=\,\menolite{}).
  (a)~Verbose generation prompts (each system's default).
  (b)~Terse prompts forcing a single direct answer; HippoRAG\,2 is
  unchanged across panels as its default already produces terse output.
  NaiveRAG is \ragu's \texttt{NaiveSearchEngine} and shares our generation
  prompt. Best AC per column in \textbf{bold}.}
\label{tab:multihop}
\setlength{\tabcolsep}{3pt}
\resizebox{\columnwidth}{!}{%
\begin{tabular}{l cc cc cc}
\toprule
& \multicolumn{2}{c}{BioASQ}
& \multicolumn{2}{c}{MuSiQue}
& \multicolumn{2}{c}{2WikiMultiHop} \\
\cmidrule(lr){2-3}\cmidrule(lr){4-5}\cmidrule(lr){6-7}
System & AC & RL & AC & RL & AC & RL \\
\midrule
\multicolumn{7}{l}{\textit{(a) Verbose generation prompt}} \\
NaiveRAG          & 55.3 & 12.4 & 41.7 & 6.9  & 43.5 & 12.0 \\
LightRAG (GPT)    & 43.9 & 6.5  & 34.5 & 3.8  & 36.2 & 8.2  \\
HippoRAG\,2 (GPT) & \textbf{74.1} & 48.8 & \textbf{56.3} & 42.7 & \textbf{65.9} & 54.7 \\
\ragu{} (GPT)     & 56.0 & 12.2 & 43.5 & 7.6  & 46.6 & 13.2 \\
\ragu{} (Ours)    & 54.5 & 12.1 & 42.0 & 7.4  & 45.2 & 12.9 \\
\midrule
\multicolumn{7}{l}{\textit{(b) Terse generation prompt}} \\
NaiveRAG          & 71.7 & 49.2 & 36.6 & 24.7 & 53.7 & 45.2 \\
LightRAG (GPT)    & 63.9 & 42.9 & 26.0 & 12.1 & 44.3 & 35.7 \\
HippoRAG\,2 (GPT) & 72.4 & 48.8 & \textbf{54.4} & 42.6 & \textbf{63.5} & 54.7 \\
\ragu{} (GPT)     & \textbf{72.9} & 48.7 & 40.1 & 26.5 & 58.0 & 49.6 \\
\ragu{} (Ours)    & 72.8 & 48.2 & 40.7 & 27.5 & 55.1 & 46.3 \\
\bottomrule
\end{tabular}%
}
\end{table}

These benchmarks are pure factoid QA with short gold answers, for which
answer \emph{format} strongly affects overlap-based metrics. We therefore
report two answer-generation protocols (Table~\ref{tab:multihop}):
(a)~each system's default verbose prompt, and (b)~a terse prompt that
forces a single direct answer. HippoRAG\,2 appears in both panels
unchanged because its default prompt already emits terse output, making
it a format anchor; NaiveRAG is \ragu's \texttt{NaiveSearchEngine} and
therefore shares our generation prompt.

Under the verbose protocol~(a), HippoRAG\,2 dominates every column (AC up
to 74.1 vs.\ 56.0 for \ragu{} on BioASQ). This gap, however, is largely a
format artifact: verbose answers mismatch the terse gold references,
depressing both ROUGE-L (12 vs.\ 49) and Answer Correctness. Once format
is controlled~(b), the picture changes substantially. \ragu{} ties and
slightly exceeds HippoRAG\,2 on BioASQ AC (72.9 vs.\ 72.4) and closes
the 2WikiMultiHopQA gap from $-$19.3\,pp to $-$5.5\,pp (58.0 vs.\ 63.5).
HippoRAG\,2 retains a genuine lead only on MuSiQue (54.4 vs.\ 40.1), the
hardest multi-hop benchmark, where its personalized PageRank follows
reasoning chains that consolidated retrieval does not surface---and where
terseness even \emph{hurts} the other systems' Answer Correctness.

The refined picture is one of complementary strengths rather than outright
dominance: HippoRAG\,2 excels at chain-following multi-hop and
single-fact precision, while \ragu{} is competitive on factoid QA once
format is controlled and superior on context-breadth tasks. Notably,
\ragu{} achieves this with a locally served 7\,B extraction model
(\menolite{}) versus HippoRAG\,2's larger 20\,B gpt-oss-20b; the gap
between \ragu{} with GPT and with \menolite{} is minimal (1--2\,pp),
confirming \menolite{} as a drop-in replacement.

\subsection{Model Evaluation}
\label{sec:meno_eval}

\begin{table}[!t]
\centering
\small
\caption{IE benchmark (knowledge-graph construction).
  NER\,=\,entity recognition (F1), RE\,=\,relation extraction (F1),
  Def\,=\,entity definition (chrF++), RDef\,=\,relation definition
  (chrF++), HM\,=\,harmonic mean of all four tasks.}
\label{tab:nerel}
\setlength{\tabcolsep}{3.5pt}
\begin{tabular}{l c cccc c}
\toprule
Model & Size & NER & Def & RE & RDef & HM \\
\midrule
\menolite{}  & 7B  & 0.504 & 0.527 & \textbf{0.347} & 0.558 & \textbf{0.468} \\
Qwen2.5-32B  & 32B & \textbf{0.536} & \textbf{0.528} & 0.239 & \textbf{0.599} & 0.416 \\
gemma-3-27b  & 27B & 0.544 & 0.482 & 0.224 & 0.583 & 0.396 \\
Qwen2.5-14B  & 14B & 0.510 & 0.518 & 0.222 & 0.583 & 0.396 \\
Qwen2.5-7B   & 7B  & 0.477 & 0.479 & 0.192 & 0.541 & 0.356 \\
T-lite-1.0   & 7B  & 0.466 & 0.464 & 0.174 & 0.533 & 0.336 \\
\bottomrule
\end{tabular}
\end{table}

Table~\ref{tab:nerel} confirms the prediction: \menolite{} achieves the
highest harmonic mean on our IE benchmark~\cite{bondarenko2026nerelbench}, outperforming Qwen2.5-32B by 12.5\% relative---driven
by relation extraction (F1 0.347 vs.\ 0.239), the sub-task most dependent
on language comprehension. On MERA~\cite{fenogenova2024mera}, \menolite{}
scores 0.555 overall with near-perfect passkey retrieval (0.98 at 128\,K
tokens, LIBRA~\cite{libra2025}), confirming robust long-context handling.

\paragraph{Fine-tuning payoff vs.\ pipeline robustness.}
\menolite{}'s large standalone extraction edge compresses to $\leq$1\,pp
on end-to-end GraphRAG-Bench QA---and does so in every pipeline we tested
(HippoRAG, LightRAG, and our own). This is not a failure of the
fine-tuning but evidence that graph-RAG QA quality is largely robust to
extractor choice once consolidation is present: \menolite{} contributes
32B-class extraction at 7\,B cost, while the consolidation pipeline
contributes downstream robustness, making the two artifacts complementary.

\section{Demonstration}
\label{sec:demo}

\paragraph{Case study.}
We illustrate \ragu's pipeline on a single sentence about Dennis
Ritchie, produced by the bundled example script:

\begin{quote}\small\itshape
Dennis Ritchie, the creator of the C programming language, and the
co-creator of the Unix operating system, died on October 12, 2011, at
the age of 70.\ \ His father, Alistair E.\ Ritchie, worked for many
years at Bell Laboratories in Murray Hill, New Jersey.
\end{quote}

\paragraph{Extraction.}
The two-stage extractor identifies 9 typed entities under the NEREL
schema (Table~\ref{tab:demo_entities}). Stage~2 then extracts 8
relations, all constrained to this validated entity set
(Table~\ref{tab:demo_relations}).

\begin{center}
\small
\captionof{table}{Entities extracted from the Ritchie passage.}
\label{tab:demo_entities}
\begin{tabular}{ll}
\toprule
\textbf{Entity} & \textbf{NEREL Type} \\
\midrule
Dennis Ritchie           & PERSON \\
Alistair E.\ Ritchie     & PERSON \\
C Programming Language   & PRODUCT \\
Unix Operating System    & PRODUCT \\
Bell Laboratories        & ORGANIZATION \\
October 12, 2011         & DATE \\
70                       & AGE \\
Murray Hill              & DISTRICT \\
New Jersey               & STATE\_OR\_PROV \\
\bottomrule
\end{tabular}
\end{center}

\begin{center}
\small
\captionof{table}{Relations extracted from the Ritchie passage (5 of 8 shown).}
\label{tab:demo_relations}
\resizebox{\columnwidth}{!}{%
\begin{tabular}{lll}
\toprule
\textbf{Source} & \textbf{Target} & \textbf{Relation} \\
\midrule
Dennis Ritchie       & C Programming Language & WORKS\_AS \\
Dennis Ritchie       & Unix Operating System  & WORKS\_AS \\
Dennis Ritchie       & October 12, 2011       & DATE\_OF\_DEATH \\
Alistair E.\ Ritchie & Dennis Ritchie         & PARENT\_OF \\
Bell Laboratories    & Murray Hill            & LOCATED\_IN \\
\bottomrule
\end{tabular}}
\end{center}

\begin{figure}[t]
\centering
\includegraphics[width=0.85\columnwidth]{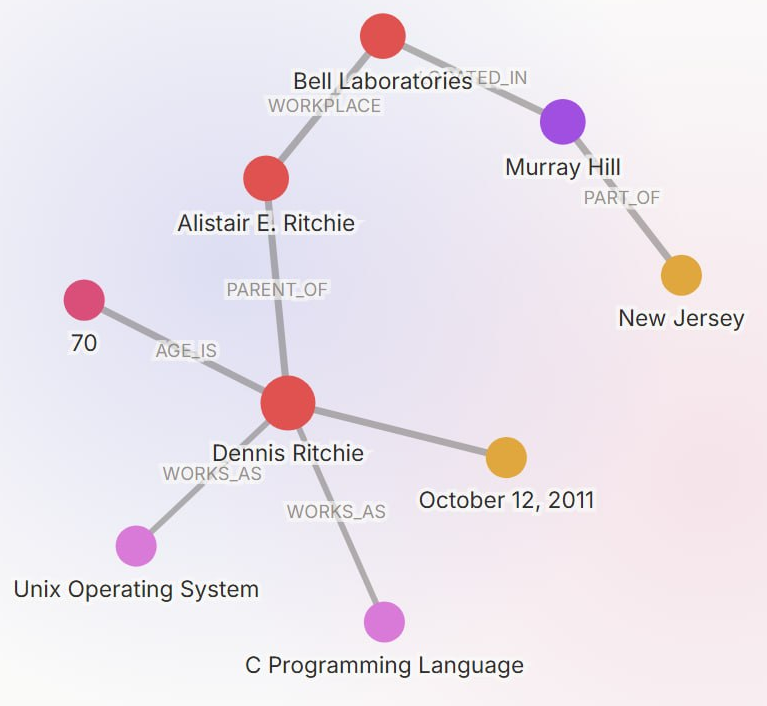}
\caption{Knowledge graph built from the Ritchie passage.
  Nodes are typed entities; edges are typed relations. Two
  communities emerge: Ritchie's professional legacy and the Bell Labs
  geographic cluster.}
\label{fig:ritchie_kg}
\end{figure}

\paragraph{Community detection.}
Leiden clustering partitions the 9-entity graph
(Figure~\ref{fig:ritchie_kg}) into two communities: one centred on
Dennis Ritchie and his creations (5~entities, 4~relations), the other
linking Bell Laboratories, Murray Hill, New Jersey, and Alistair Ritchie
through spatial and professional ties (4~entities, 3~relations). An LLM
generates a structured summary for each.

\paragraph{Multi-hop retrieval.}
Using the built graph, \ragu's \texttt{LocalSearchEngine} answers
questions that require traversing multiple edges:
\begin{quote}\small
\textbf{Q:} Where did the father of the creator of the C programming
language work? \\[2pt]
\textbf{A:} Alistair E.\ Ritchie\ldots\ worked at
\textbf{Bell Laboratories}. \\[4pt]
\textbf{Q:} What did the person who died on October 12, 2011, create? \\[2pt]
\textbf{A:} Dennis Ritchie\ldots\ created the
\textbf{C programming language} and co-created the
\textbf{Unix operating system}.
\end{quote}

\paragraph{Availability.}
\ragu{} is installable via \texttt{pip install graph\_ragu}; full API
documentation and examples are at
\url{https://github.com/RaguTeam/RAGU}.
A demonstration video is available at
\url{https://youtu.be/bicJDMJuQfg}.
The \ragu{} system is released under the MIT license (code at
\url{https://github.com/RaguTeam/RAGU}), and the \menolite{} model
is distributed under Apache\,2.0 at
\url{https://huggingface.co/bond005/meno-lite-0.1}.

\section{Conclusion}
\label{sec:conclusion}

We argued that the LLM inside a RAG pipeline needs language skills---not
world knowledge---and that these skills scale weakly with model size.
\ragu{} operationalizes this insight via a modular multi-step pipeline
that retrieves the most complete context at every factoid level of
GraphRAG-Bench and overtakes HippoRAG\,2 on synthesis tasks (Creative
Generation AC and Coverage); HippoRAG\,2 conversely excels at retrieval
precision---leading single-fact AC and chain-following multi-hop
reasoning on MuSiQue. The wider multi-hop gap seen under verbose prompts
is largely an answer-format artifact. Practically: prefer \ragu{} when
answers must synthesize broad context (summarization, creative
generation, long-form QA) under a single-GPU budget, and prefer
chain-traversal systems for precise multi-hop fact lookup.
Both artifacts are released under open-source licenses.

\section*{Limitations}

First, our scaling evidence rests on a single model family (Qwen2.5) and
select tasks; although robust across six model sizes, it is a
well-supported hypothesis rather than a universal theorem.

Second, \menolite{} trades parametric factual recall for contextual
grounding and should not serve as a standalone knowledge base; its
multi-hop reasoning degrades beyond 32K tokens, typical of 7\,B-class
models. We also note a distributional-overlap caveat on our IE benchmark
~\cite{bondarenko2026nerelbench}: the supervised fine-tuning of \menolite{}
uses the train and validation splits of NEREL, whereas the benchmark
uses only the held-out test split with different instruction wordings;
the overlap is confined to the annotation schema and text domain, not to
the benchmark documents---but a residual advantage cannot be fully ruled
out.

Finally, \ragu{}'s default NetworkX graph backend---a swappable
\texttt{BaseGraphStorage} adapter---does not by itself scale to massive
corpora (millions of nodes), so such deployments require a dedicated
graph-database adapter. Final graph quality remains sensitive to the
extraction LLM: weak base models introduce structural noise that
consolidation cannot fully rectify.

\section*{Ethics Statement}

\paragraph{Data Provenance and Licensing.}
The \ragu{} system is released under the MIT
license, while the \menolite{} model is distributed under Apache\,2.0.
\menolite{} was trained exclusively on publicly available datasets that
are cited in the References---including educational web
corpora~(FineWeb-Edu), Russian-language academic texts~(RuLM),
information-extraction datasets~\cite{loukachevitch2021nerel},
multi-hop QA benchmarks~\cite{tang2024multihop,Katsis2025mtrag}, and
synthetic instructions generated with GPT-4o-mini. No personally
identifiable information was included in any training or evaluation
corpus. The IE benchmark is a test-only derivative of the human-annotated
NEREL corpus~\cite{nerel-instruct} and is released under an MIT license; its integration into
the LM Evaluation Harness~\cite{evalharness2024} enables standardized,
reproducible evaluation.

\paragraph{Environmental Impact and Democratization.}
A central design goal of this work is to make high-quality GraphRAG
accessible beyond large industrial laboratories. Because the
language-knowledge hypothesis (\S\ref{sec:intro}) predicts---and our
experiments confirm---that a 7\,B model suffices for extraction,
\ragu{}+\menolite{} runs on a \emph{single consumer GPU}, rather than
the multi-GPU clusters required by frontier models. This is not an
aspirational claim but a measured consequence: graph construction
processes $\sim$8\,k tokens/document at $\sim$2\,k tok/s, costing
$\sim$\$0.001/doc on rented GPUs---roughly two orders of magnitude less
than the $\sim$\$0.10/doc of commercial API-based alternatives
(Appendix~\ref{app:cost}). At corpus scale (100\,k documents), the
difference is $\sim$\$100 (local GPU) vs.\ $\sim$\$10\,000 (commercial
API). The lower compute footprint translates directly into lower energy
consumption and CO\textsubscript{2} emissions per document indexed.
Furthermore, by supporting both English and Russian through
\texttt{Settings.language} and by training on Russian-language corpora,
\ragu{} lowers the barrier for non-English-speaking users and
underrepresented language communities.

\paragraph{Potential Misuse.}
Like any RAG system, \ragu{} can amplify biases present in its indexing
corpus: if the source documents contain prejudiced or factually incorrect
content, the extracted knowledge graph will reflect those biases, and
generated answers may inherit them. \menolite{} additionally inherits
biases from its pretraining and fine-tuning corpora. We recommend
domain-specific evaluation before deployment in sensitive applications
(healthcare, legal, news). \menolite{} deliberately trades parametric
factual recall for context-grounded skills and should not be used as a
standalone knowledge base, reducing the risk of confident hallucination
but increasing dependence on corpus quality. On the engineering side,
\ragu{} validates all LLM outputs through Pydantic models rather than
executing raw model responses (unlike systems that use \texttt{eval()}),
which eliminates a class of code-injection attacks from adversarial
model output.

\paragraph{Bias and Fairness.}
The NEREL schema~\cite{loukachevitch2021nerel} underlying \ragu's
extraction was developed for Russian news text; its entity and relation
types reflect that domain. Applying \ragu{} to other languages or domains
may require schema adaptation. The IE benchmark similarly reflects
Russian-language text characteristics (e.g., heavy inflection, addressed
by Snowball stemming in the metric). Users should be aware of potential
performance degradation when operating outside the schema's design
domain.

\paragraph{Transparency and Reproducibility.}
All code, model weights, and benchmark data are publicly released under
open licenses. The repository ships $\sim$374 automated tests and a
deterministic mock LLM server, enabling full regression testing without
API keys or GPU resources---a property that, to our knowledge, no other
open-source GraphRAG framework provides. Every domain object (entity,
relation, chunk) carries a deterministic MD5 identifier, allowing any
retrieved result to be traced back to its source text. The IE benchmark
is integrated into the public LM Evaluation Harness repository under
\nereltask{}, ensuring that any causal LLM can be evaluated under
identical conditions.

\section*{Acknowledgments}

RAGU development was supported by GitVerse, Cloud.ru, and Habr through
the ``Code Without Borders'' open-source grant program (first place, AI
Innovation)\footnote{\url{https://habr.com/ru/specials/979702/}} and by
Yandex through the Yandex Open Source grant program (Mikhail Komarov,
first place, Artificial Intelligence
track).\footnote{\url{https://habr.com/ru/companies/yandex/articles/1040282/}}
We thank Kirill~Novgorodtsev for the web frontend (built with the
\texttt{\$mol} framework)\footnote{\url{https://github.com/hyoo-ru/mam_mol}}
and Yaroslav~Svetlov for the backend of the RAGU demo
website.\footnote{\url{https://raguteam.github.io/web/\#!api=https\%3A\%2F\%2Fragu-back.duckdns.org}}

\bibliography{references}

\appendix

\section{Engineering Comparison}
\label{app:engineering}

Industrial adoption of GraphRAG is constrained less by retrieval quality
than by the engineering discipline of available implementations.
Table~\ref{tab:eng_comparison} compares \ragu{} with HippoRAG\,2 along
the production risks we encountered while benchmarking open-source
GraphRAG frameworks.\footnote{We focus on HippoRAG\,2 as the most
  architecturally comparable system---recent, academic, actively
  maintained; similar observations apply to several other frameworks in
  our survey. All file:line references in this appendix point to commit
  \texttt{d437bfb1} of \texttt{OSU-NLP-Group/HippoRAG} (2025-09-04, HEAD
  of \texttt{main} at the time of analysis; package version
  \texttt{hipporag\,2.0.0-alpha.4}), so the comparison is reproducible
  against a fixed snapshot rather than a moving target.}
Both systems are open-source and support incremental indexing; the
differences lie in how each handles failure, migration, and change.

\begin{table}[!htbp]
\centering
\footnotesize
\caption{Engineering comparison between \ragu{} and HippoRAG\,2,
  organized by the production risk each property addresses.}
\label{tab:eng_comparison}
\setlength{\tabcolsep}{4pt}
\renewcommand{\arraystretch}{1.05}
\begin{tabularx}{\linewidth}{@{} >{\raggedright\arraybackslash}X >{\raggedright\arraybackslash}X @{}}
\toprule
\textbf{\ragu{}} & \textbf{HippoRAG\,2} \\
\midrule
\multicolumn{2}{@{}l}{\textit{Silent data loss on crash}} \\
Lifecycle callbacks at every storage tier & No shared flush protocol \\
\multicolumn{2}{@{}l}{\textit{Backend migration cost}} \\
Three swappable tiers behind one interface & Concrete Parquet/Pickle/SQLite; pipeline rewrite required \\
\multicolumn{2}{@{}l}{\textit{API-bound throughput}} \\
Native async with semaphore-bounded batching and rate limits & Synchronous core; thread-pool parallelism \\
\multicolumn{2}{@{}l}{\textit{Code execution from LLM output}} \\
Pydantic-validated structured outputs & \texttt{eval()} on raw LLM responses \\
\multicolumn{2}{@{}l}{\textit{Transient failure recovery}} \\
\texttt{tenacity} retry; per-chunk error isolation & Broad \texttt{except}; \texttt{assert False} as control flow \\
\multicolumn{2}{@{}l}{\textit{Regression detection}} \\
$\sim$374 tests; deterministic mock LLM server & Demo scripts requiring live API keys; no \texttt{pytest}, no CI \\
\multicolumn{2}{@{}l}{\textit{Incremental maintenance}} \\
Explicit upsert/update/delete; merge policies; consistency auditor & Hash dedup; file-existence checks \\
\multicolumn{2}{@{}l}{\textit{Reproducible deployment}} \\
Loose version constraints; optional GPU extras & Hard-pinned \texttt{torch}+\texttt{vllm}; CUDA-locked \\
\multicolumn{2}{@{}l}{\textit{Modularity}} \\
Thirteen abstract base classes; constructor injection & Monolithic indexing class ($\sim$1.6k lines) \\
\bottomrule
\end{tabularx}
\end{table}

Two entries deserve emphasis because they shape failure modes in
production. HippoRAG\,2 parses model output through Python's
\texttt{eval()} applied to a regex-filtered substring of the raw LLM
response---an arbitrary-code-execution surface should the model emit
hostile content, and a source of opaque exceptions far from the call
site on any syntactic deviation. It also relies on \texttt{assert} for
control flow: the offline indexing path terminates with
\texttt{assert False} (\texttt{HippoRAG.py:216}), so under
\texttt{python -O} the assertion is stripped and the offline path
silently proceeds into indexing without the online vLLM server the rest
of the pipeline expects.

The cumulative effect is felt at migration time.
Moving \ragu{} from a single-machine prototype (NetworkX, NanoVDB) to a
production stack (Neo4j, Qdrant with hybrid retrieval) requires changing
two constructor arguments; the same migration in a system without
storage abstractions requires reimplementing the indexing pipeline.
Likewise, the mock LLM server reduces a full CI run from dollars of API
calls to seconds of CPU time, which is what makes continuous regression
testing affordable on a GraphRAG codebase at all.

\paragraph{Provenance.}
Every HippoRAG\,2 claim is verifiable against the pinned commit. Key
anchors (paths relative to repository root):
\texttt{eval()} on raw LLM output at
\href{\hipporoot src/hipporag/information_extraction/openie_openai.py#L36}{\nolinkurl{openie\_openai.py:36,88}};
\texttt{assert False} at
\href{\hipporoot src/hipporag/HippoRAG.py#L216}{\nolinkurl{HippoRAG.py:216}}
and
\href{\hipporoot src/hipporag/embedding_model/__init__.py#L30}{\nolinkurl{embedding\_model/\_\_init\_\_.py:30}};
bare \texttt{except} at
\href{\hipporoot src/hipporag/information_extraction/openie_openai.py#L63}{\nolinkurl{openie\_openai.py:63,112}};
monolithic \texttt{class HippoRAG} (1611 lines) at
\href{\hipporoot src/hipporag/HippoRAG.py}{\nolinkurl{HippoRAG.py}};
concrete storage backends (Parquet at
\href{\hipporoot src/hipporag/embedding_store.py#L40}{\nolinkurl{embedding\_store.py:40}},
Pickle at \href{\hipporoot src/hipporag/HippoRAG.py#L183}{\nolinkurl{HippoRAG.py:183}},
SQLite at \href{\hipporoot src/hipporag/llm/transformers_llm.py#L37}{\nolinkurl{transformers\_llm.py:37}});
no shared flush between
\texttt{EmbeddingStore.\_save\_data} and
\texttt{HippoRAG.save\_igraph}
(\href{\hipporoot src/hipporag/HippoRAG.py#L1088}{\nolinkurl{HippoRAG.py:1088}});
no \texttt{pytest} or \texttt{.github/} in the repository;
\href{\hipporoot requirements.txt#L3}{\nolinkurl{requirements.txt:3,5}}
hard-pins \texttt{vllm} and \texttt{torch}. These choices are reasonable
for reproducing the HippoRAG\,2 paper's own experiments; the trade-offs
above matter mainly outside that setting.

\section{GraphRAG-Bench Ablation Summary}
\label{app:ablation}

Table~\ref{tab:ablation} reports generation AC for selected \ragu{}
configurations. Full generation and retrieval results for all 11
configurations are in the repository.

\begin{center}
\small
\captionof{table}{Generation AC ($\times$100) for selected \ragu{}
  configurations on GraphRAG-Bench Medical (ICL\,=\,1, Val\,=\,yes
  unless noted).}
\label{tab:ablation}
\setlength{\tabcolsep}{3pt}
\begin{tabular}{cc l ccc c}
\toprule
ICL & Val & LLM & FR & CR & CS & CG \\
\midrule
0 & no & Qwen2.5-3B  & 53.1 & 53.1 & 63.8 & 58.2 \\
0 & no & Qwen2.5-7B  & 54.3 & 53.4 & 64.1 & 58.2 \\
0 & no & \menolite{}  & 54.3 & 54.0 & 63.9 & 58.8 \\
\midrule
1 & no & Qwen2.5-14B & 53.9 & 54.2 & 65.6 & 59.1 \\
1 & yes & Qwen2.5-7B & 54.1 & 54.6 & 64.9 & 58.1 \\
1 & yes & \menolite{} & 54.2 & 53.7 & 64.1 & 59.0 \\
\bottomrule
\end{tabular}
\end{center}

Key observations: model size (3B--14B) shifts AC by $\leq$1.5\,pp;
ICL and validation each contribute $<$1\,pp;
\menolite{} and Qwen2.5-7B are within 1\,pp in every configuration.

\section{Cost Analysis}
\label{app:cost}

Table~\ref{tab:cost_detail} reports graph-construction (indexing) cost, a
one-time per-document operation, separate from answer generation
(query-time, recurring). All systems share the same answer-generation
model (gpt-4o-mini via API), so query-time cost is a common baseline.
For construction, MS-GraphRAG uses a commercial gpt-4o API (priced at
\$2.50/M input tokens~\cite{openai_pricing}); the remaining systems run
a local model under vLLM~\cite{kwon2023efficient}---LightRAG and
HippoRAG\,2 use gpt-oss-20b, and \ragu{} uses \menolite{}---billed at a
fixed GPU-hour rate ($\sim$\$0.001/doc on rented GPUs at $\sim$\$1/h
and $\sim$2\,k tok/s; near zero on owned hardware). At corpus scale
(100\,k documents), construction cost is $\sim$\$10\,000 for MS-GraphRAG
vs.\ roughly \$100 for \ragu{}+\menolite{}, with LightRAG and
HippoRAG\,2 in the same GPU-cost class. Token volumes are averaged
empirical measurements; the MS-GraphRAG figure reflects the token
intensity of its global indexing approach~\cite{edge2024graphrag}.
Note that volumes are computed with each model's own tokenizer and are
not directly comparable across systems.

\begin{center}
\small
\captionof{table}{Approximate graph-construction (indexing) token volume
  and cost per document; query-time answer generation (gpt-4o-mini, common
  to all systems) is omitted. $^{*}$ marks empirical measurements from our
  experiments; the unmarked MS-GraphRAG count is an order-of-magnitude
  estimate.}
\label{tab:cost_detail}
\setlength{\tabcolsep}{3pt}
\resizebox{\columnwidth}{!}{%
\begin{tabular}{l lc l}
\toprule
System & Indexing model & Tokens/doc & Cost/doc \\
\midrule
MS-GraphRAG (global) & gpt-4o (API)        & $\sim$40\,k        & $\sim$\$0.10 \\
HippoRAG\,2          & gpt-oss-20b (local) & $\sim$6\,k$^{*}$   & fixed GPU \\
LightRAG             & gpt-oss-20b (local) & $\sim$8\,k$^{*}$   & fixed GPU \\
\ragu{}+\menolite{}  & \menolite{} (local) & $\sim$8\,k$^{*}$   & fixed GPU \\
\bottomrule
\end{tabular}%
}
\end{center}

\end{document}